\documentclass[10pt,conference,letterpaper]{IEEEtran}
\IEEEoverridecommandlockouts
\usepackage{cite}
\usepackage{amsmath,amssymb,amsfonts}
\usepackage[linesnumbered,ruled,vlined]{algorithm2e}
\usepackage{algpseudocode}
\usepackage{graphicx}
\usepackage{textcomp}
\usepackage{xcolor}
\usepackage{xspace}
\usepackage{url}
\usepackage{hyperref}
\usepackage[capitalize]{cleveref}
\usepackage{float}
\usepackage{subcaption}
\usepackage{multicol}
\usepackage{multirow}
\usepackage{wrapfig}
\usepackage[shortlabels]{enumitem}
\usepackage{balance}
\usepackage{pifont}

\def\BibTeX{{\rm B\kern-.05em{\sc i\kern-.025em b}\kern-.08em
    T\kern-.1667em\lower.7ex\hbox{E}\kern-.125emX}}

\newcommand{\proj}{\emph{UniPipe}\xspace}

\begin{document}

\title{\proj: Unified Inference Serving and Fine-Tuning for Deadline-Constrained DNN Surrogate Workflows}

% \author{
% \IEEEauthorblockN{Jaiaid Mobin\textsuperscript{*}, Avinash Maurya\textsuperscript{\dag}, M.~Mustafa Rafique\textsuperscript{\ddag}, Bogdan Nicolae\textsuperscript{\S}}
% \IEEEauthorblockA{\textit{\textsuperscript{*}jm5071@rit.edu,\textsuperscript{\ddag}mrafique@cs.rit.edu,\{\textsuperscript{\dag}amaurya,\textsuperscript{\S}bnicolae\}@anl.gov}\\\textit{\textsuperscript{*\ddag}Rochester Institute of Technology, USA; \textsuperscript{\dag\S}Argonne National Laboratory, USA}}
% }

\author{
    \IEEEauthorblockN{
        Jaiaid Mobin\IEEEauthorrefmark{1},
        Avinash Maurya\IEEEauthorrefmark{2},        
        M. Mustafa Rafique\IEEEauthorrefmark{1},
        Bogdan Nicolae\IEEEauthorrefmark{2},
    }
    \IEEEauthorblockA{
        \IEEEauthorrefmark{1}\it{Rochester Institute of Technology, USA};
        \IEEEauthorrefmark{2}\it{Argonne National Laboratory, USA}\\
        Email: {\IEEEauthorrefmark{1}\{jm5071@rit.edu, mrafique@cs.rit.edu\}; \IEEEauthorrefmark{2}\{amaurya, bnicolae\}@anl.gov;}  
    }
}

\newcommand{\arxivheader}{%
  \smash{\raisebox{6.25cm}{\makebox[\textwidth][c]{\small This work has been accepted for publication in the 2026 IEEE International Conference on e-Science (eScience).}}}%
}
\IEEEpubid{\copyright~2026 IEEE. Personal use of this material is permitted. Permission from IEEE must be obtained for all other uses.}
\maketitle
\arxivheader

\begin{abstract}
Deep neural network (DNN) surrogates are increasingly used to accelerate scientific workflows that must process high-rate sensor streams in near real time, exemplified by Ptychographic imaging at next-generation X-ray light sources. Such surrogates deliver $100$--$1000\times$ higher throughput than classical iterative solvers, enabling faster steering of experiments. To preserve surrogate accuracy during an experiment, as the input distribution evolves, state-of-the-art deployments either freeze the pretrained surrogate or run a separate CUDA context that fine-tunes concurrently with serving. However, multi-context concurrency results in significant contention on GPU execution, inflating both the mean and the variance of inference latency, triggering deadline violations in workflows where sensor data (e.g., image, diffraction pattern) has to be processed within a hard per-unit deadline. To address this challenge, we propose \proj, a single-context system that unifies inference serving and fine-tuning of a DNN surrogate on a single edge GPU. \proj is driven by three key insights: both serving and fine-tuning require a forward pass and can be fused into one; the per-sample latency of modern GPUs is non-linear in batch size; and the workload is repetitive across intervals, enabling offline scheduling. Specifically, \proj fuses training and inference data in a single forward pass with a selective backward pass on the training subset, and uses a polynomial-time scheduler that exploits the non-linear latency curve to compute the batch composition under the deadline. We evaluate \proj on a real-world Ptychographic reconstruction workflow with sensor rates up to $5000$ patterns/sec.; \proj lowers reconstruction MSE by up to $30\%$ and inference MSE by up to $50\%$ over a pretrained baseline, and reduces the deadline-miss rate by $13\%-22\%$ over a multi-context baseline. \proj's unified deployment reduces the latency spread from $106\%$ to $57\%$ of the mean, and demonstrates higher total throughput (training + inference) compared to multi-context baseline.
\end{abstract}


\section{Introduction}
\label{section:intro}
%==================================================================
A growing number of scientific workflows pair a high-throughput sensor with a downstream reconstruction step that must complete in near real time to steer the experiment, focus on quantities of interest, or trigger early termination of unproductive runs. Examples include coherent diffractive imaging~\cite{maiden2009improved,babu2023deep}, computational fluid dynamics~\cite{chen2021numerical}, and neural PDE solvers~\cite{brandstetter2022message}. Ptychography, a specialized form of coherent diffractive imaging that uses many spatially overlapped diffraction patterns to recover sub-angstrom 
\IEEEpubidadjcol
features~\cite{babu2023ai,jiang2018electron}, is a particularly demanding case: next-generation beamlines such as APS-U~\cite{kerby2023advanced} are up to $500\times$ brighter than their predecessors and produce raw diffraction streams of $300$~Gbps~\cite{gao2025x}. Classical iterative reconstruction tools, such as Tike~\cite{tike}, recover the underlying phase images with high fidelity but cannot keep up with sensor data rates~\cite{gao2025x}. While DNN surrogates, such as PtychoNN~\cite{cherukara2020ai}, close this throughput gap by $100$--$1000\times$~\cite{cherukara2020ai,babu2023deep,nakahata2024ptychoformer} at the price of reduced accuracy that further drifts as the features being scanned change.

\paragraph*{\bf Motivation: Continual Fine-Tuning under Hard Deadlines}
On a representative Ptychographic stream, freezing a pretrained PtychoNN surrogate leaves a persistent error gap to the oracle that widens over regions whose features were under-represented at offline training time, while continual fine-tuning on freshly labeled samples closes most of this gap (\cref{fig:concept_drift_demonstration}). A typical way to obtain continual fine-tuning, i.e., running a second model instance in a separate CUDA context that trains concurrently with the serving instance, is, however, fatally expensive for deadline-constrained serving. On our edge testbed, evaluation shows multi-context concurrency inflates the per-sample inference latency by $\sim\!2\times$ in the mean and also shows significant fluctuations (\cref{fig:single_multicontext_benchmark}). This variability is particularly damaging because a deadline-aware scheduler must respect the worst case, forcing conservative batching that further reduces useful throughput and, at the data rates expected for next-generation beamlines, triggers a $13\%-22\%$ increase in deadline-miss rate compared with serving-only deployments.

\paragraph*{\bf Limitations of State-of-the-Art}
Existing approaches such as periodic-halt fine-tuning~\cite{babu2023deep,lexelius2025ptychographic}, multi-context concurrent fine-tuning~\cite{wang2019adaptive,wortsman2022model,houlsby2019parameter}, and remote-HPC offloaded fine-tuning~\cite{babu2023ai,cherukara2022_hpcai_xray} suffer from two specific limitations. First, those that train and serve in distinct CUDA contexts contend for streaming multiprocessors, memory bandwidth, and the kernel-launch queue, producing the latency variability shown in \cref{fig:single_multicontext_benchmark} and discussed above. Second, those that statically partition the GPU, e.g., via NVIDIA's Multi-Process Service (MPS)~\cite{cuda_mps,cuda_mps_oakridge}, cannot recover the lost latency stability because contention also occurs at the memory subsystem and kernel-launch path, neither of which is partitioned by MPS. \textit{None of these approaches exploits the fact that inference and fine-tuning share a forward pass and can therefore be fused into a single CUDA context, eliminating cross-context contention at its source.}

\paragraph*{\bf Key Insights and Contributions}
We leverage \textbf{three key insights}: (i)~both inference and fine-tuning require a forward pass, and only fine-tuning additionally requires a backward pass, so the forward pass can be fused across a mixed batch of unlabeled inference requests and labeled training samples; (ii)~the per-sample latency of modern GPUs is non-linear in batch size, with a flat regime below the saturation knee where additional samples are essentially free; and (iii)~the workload is repetitive across fixed-length intervals, so the schedule can be computed once, offline. Driven by these insights, we propose \proj, a single-context system that unifies inference serving and fine-tuning of a DNN surrogate on a single edge GPU under a hard per-request deadline. Specifically, the contributions of this work are:

\begin{enumerate}[leftmargin=*,label=(\arabic*)]
    \item We perform an in-depth characterization of the accuracy-latency trade-off of online fine-tuning for DNN surrogates on edge GPUs, quantifying the latency-stability cost of multi-context CUDA execution and motivating a single-context design~(\S\ref{section:motivation}).
    
    \item We propose a set of design principles for unified, single-context inference and fine-tuning, comprising overlapped classical and DNN scheduling, a fused forward pass with selective backward pass on the labeled subset, and exploitation of the non-linear per-sample latency curve of modern GPUs~(\S\ref{subsection:sysdesign_design_principle}).
    
    \item We design a polynomial-time dynamic-programming scheduler that computes the batch composition under the per-unit deadline, and implement \proj as a three-process system integrated with the PtychoNN surrogate~(\S\ref{sec:algo},~\S\ref{subsection:unipipe_implementation}).

    \item We evaluate \proj on a real-world Ptychographic reconstruction workflow with PtychoNN: it reduces inference MSE by up to $50\%$ and reconstruction MSE by up to $30\%$ over a pretrained baseline, and lowers the deadline-miss rate by $13\%$--$22\%$  while shrinking the $95^{\text{th}}$-percentile latency spread from $106\%$ to $57\%$ of the mean over a multi-context baseline~(\S\ref{section:evaluation}).
\end{enumerate}

\paragraph*{\bf Limitations of the Proposed Approach}
\proj is designed for a single-device edge node and does not address multi-GPU scaling. The dynamic-programming scheduler assumes that the forward and backward latency curves are profileable once before deployment and stay completely stable in deployment. \proj also assumes that the surrogate's forward pass is mode-invariant; surrogate architectures with mode-dependent layers, e.g., BatchNorm, Dropout, require either per-mode profiling or running the labeled and unlabeled sub-batches separately. Finally, \proj assumes that the classical-solver throughput $c_{rate}$ is stable; drift in $c_{rate}$ would require re-running the scheduler, which may be costlier for smaller surrogates where batch sizes can be larger. \proj has structural requirements--mode-invariant forward pass, profileable latency curve, repetitive intervals, and \proj is evaluated only on one workflow. Therefore, this work demonstrates empirical generality.
% =================================================
% =================================================
% =================================================

\section{Problem Description}
\label{section:prob_formulation}

\begin{figure}
    \centering
    \includegraphics[width=\linewidth]{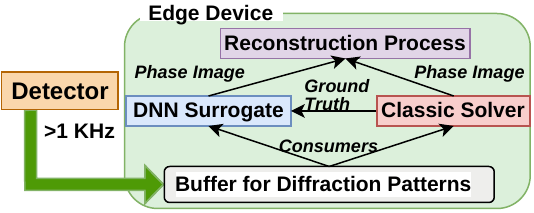}
    \caption{DNN surrogate workflow for Ptychographic reconstruction at the edge. A detector produces diffraction patterns at rate $d_{rate}$; an edge node must reconstruct each pattern within a per-pattern deadline $D$ using either a fast DNN surrogate (lower accuracy) or a slow classical solver (higher accuracy).}
    \label{fig:problem_scenario} 
\end{figure}

We consider the workflow shown in \cref{fig:problem_scenario}, which abstracts the real-time Ptychographic imaging setup described in~\cite{babu2023deep,lexelius2025ptychographic,cherukara2022_hpcai_xray,ptychonn_edge_setup}. A sensor (producer) emits data units--in our running example, diffraction patterns--at a fixed rate $d_{rate}$. Each data unit must be processed within a hard per-unit deadline $D$, measured from the unit's arrival. The processed outputs (in our example, phase images) feed a downstream reconstruction step that assembles the final scientific artifact.

The edge node has sufficient memory and compute to run both a classical solver on CPU and a DNN surrogate on a single GPU. We may route each unit either to the classical solver, which produces a high-fidelity output but at an aggregate rate $c_{rate} \ll d_{rate}$, or to the DNN surrogate, which produces an approximate output at a much higher rate. Outputs produced by the classical solver are also retained as ground truth and are used to continuously fine-tune the DNN surrogate. If a unit's deadline expires before its output is ready, either by the classical solver or by the surrogate, a proxy function $g(\cdot)$ (in our setting, interpolation from neighboring reconstructed patches) supplies a fallback estimate.

Let $\hat{\phi}_i$ denote the produced output for the $i$-th unit and $\phi_i^*$ its ground-truth counterpart. Let $F(b)$ and $B(b)$ denote, respectively, the forward- and backward-pass latency of the DNN surrogate for batch size $b$; both are characterized empirically once per model before deployment. The optimization goal is:
\begin{align}
\min \sum_i \mathrm{MSE}(\hat{\phi}_i, \phi_i^*) \quad
\text{s.t. } t_i^{\text{ready}} \le t_i^{\text{arrive}} + D, \forall i
\label{eq:objective}
\end{align}
where the per-unit completion time $t_i^{\text{ready}}$ depends on the (i)~routing decision (classical vs.\ DNN surrogate), (ii)~batch composition decisions taken by the GPU scheduler, and (iii)~the order in which units are admitted into batches. The deadline is a property of the workflow: an output produced after $D$ can no longer steer the experiment and is replaced by $g(\cdot)$. When $d_{rate}$ exceeds the surrogate's sustainable rate the feasible set is empty; \proj then solves the best-effort relaxation, maximizing the units served within $D$ rather than guaranteeing all of them. \textit{To address this objective}, we next survey prior approaches that target related aspects of the problem (\S\ref{section:related}) and in \S\ref{section:motivation} present the empirical observations that drive \proj's design.
% =================================================
% =================================================
% =================================================

\section{Related Work}
\label{section:related}
%==================================================================

We organize prior work into three themes that bracket \proj's design space: continual learning under resource constraints, real-time knowledge distillation, and DNN surrogates for Ptychography. We discuss each in turn.

\paragraph*{\bf Continual learning under resource constraints}
Continual-learning techniques typically guard against catastrophic forgetting by replaying samples from a rehearsal buffer~\cite{parisi2019continual,wu2026review,intro_rebuffi2017icarl,gao2023sar,CCGrid24-RehearsalBuffers,RehearsalBuff-FGCS24}. Replay is effective but expensive in memory and compute, which makes it hard to use under tight deadlines on edge hardware~\cite{parisi2019continual}. \textit{None of these approaches addresses the contention between concurrent serving and fine-tuning that arises when both run on the same edge GPU.} \proj sidesteps catastrophic forgetting by running over a stationary stream in which the classical solver continuously supplies fresh, in-distribution ground truth, and spends the saved cycles on keeping inference latency predictable.

\paragraph*{\bf Real-time knowledge distillation}
Distilling a large teacher into a small student to obtain a deployable model under tight latency is a well-studied technique~\cite{stanton2021does,mansourian2025comprehensive}. On a single edge device the teacher's cost is prohibitive, so existing approaches rely on cloud-side teachers~\cite{liu2025enabling} or aggressive model compression~\cite{paparounas2024evaluating}. \textit{Neither line of work exploits the structural overlap between the teacher's and the student's forward pass that \proj's fused-batch design relies on.} \proj does not target distillation, but its fused forward pass and scheduler could be reused in a fast-student / slow-teacher setup that updates the student in place.

\paragraph*{\bf DNN surrogates for Ptychography}
Recent work has demonstrated both supervised~\cite{nakahata2024ptychoformer,cherukara2020ai,babu2023deep,cam2025ptychographic} and unsupervised~\cite{hoidn2023physics} DNN surrogates for Ptychographic reconstruction, and real-time deployments~\cite{babu2023ai,lexelius2025ptychographic,cherukara2020ai} either freeze the surrogate or offload fine-tuning to a remote HPC tier. \textit{None of these deployments addresses the single-device case in which fine-tuning must run concurrently with deadline-constrained serving.} \proj is orthogonal to the choice of surrogate architecture: any model whose forward pass is mode-invariant and whose latency curve is profileable (P3) can plug into \cref{alg:optimal_makespan_scheduler} after a one-time profiling pass.
% =================================================
% =================================================
% =================================================

\section{Motivating Observations}
\label{section:motivation}
%==================================================================

Two empirical observations motivate \proj's design. We collected them on the testbed and workload described in \cref{sec:method}.

% \begin{figure}[t]
%     \begin{minipage}[t]{0.49\linewidth}
%         \centering
%         \includegraphics[width=\linewidth]{figures/exp_motivation/fig_oracle_20percentpretrained_intervaltrained_error_distribution.pdf}
%         \caption{Effect of online fine-tuning on inference accuracy. \textit{Lower is better.}}
%         \label{fig:concept_drift_demonstration}
%     \end{minipage}\hfill
%     \begin{minipage}[t]{0.49\linewidth}
%         \centering
%         \includegraphics[width=\linewidth]{figures/exp_motivation/fig_single_multicontext_benchmark_ptychonn.pdf}
%         \caption{Per-sample inference latency: single context vs.\ multi-context. \textit{Lower is better.}}
%         \label{fig:single_multicontext_benchmark}
%     \end{minipage}
% \end{figure}

% \begin{figure}[t]
%     \centering
%     \includegraphics[width=0.8\linewidth]{figures/exp_motivation/fig_oracle_20percentpretrained_intervaltrained_error_distribution.pdf}
%     \caption{Effect of online fine-tuning on inference accuracy. \textit{Lower is better.}}
%     \label{fig:concept_drift_demonstration}
% \end{figure}

\begin{figure}[t]
    \centering
    \begin{minipage}[t]{0.49\linewidth}
        \includegraphics[width=\linewidth]{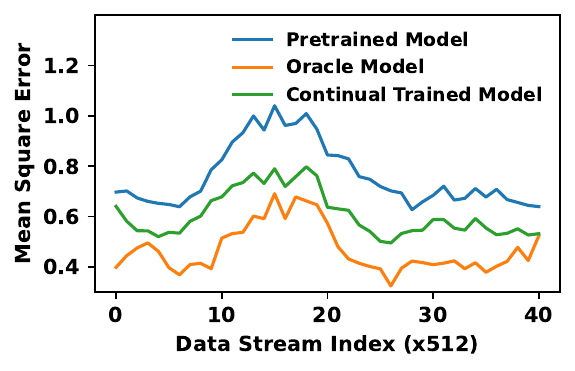}
    \caption{Effect of online fine-tuning on inference accuracy. \textit{Lower is better.}}
    \label{fig:concept_drift_demonstration}
    \end{minipage}\hfill
    \begin{minipage}[t]{0.49\linewidth}
        \includegraphics[width=\linewidth]{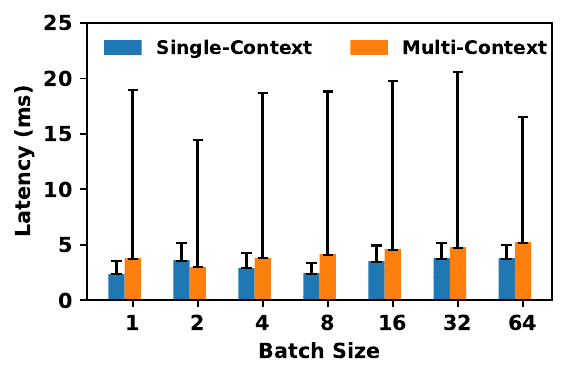}
    \caption{Per-sample inference latency: single context vs.\ multi-context. \textit{Lower is better.}}
    \label{fig:single_multicontext_benchmark}
    \end{minipage}
\end{figure}

\paragraph*{\bf Observation 1: Online fine-tuning closes the oracle gap}
We compare three configurations on the same diffraction stream: (i)~\emph{Pretrained}: a PtychoNN model trained offline on $20\%$ portion of the dataset and frozen for the live run; (ii)~\emph{Continual Trained}: the same pretrained model, fine-tuned continuously on a sparse, evenly spaced subset of incoming patterns used as ground truth; (iii)~\emph{Oracle}: a PtychoNN model trained from scratch on the full dataset, representing the upper bound that the surrogate architecture can reach. \cref{fig:concept_drift_demonstration} plots the per-pattern MSE as the experiment progresses (the $x$ axis indexes the position of the pattern in the stream). The pretrained model exhibits a persistent gap to the oracle that widens over portions of the scan whose features were underrepresented at the offline training time. Continual fine-tuning closes most of this gap, and the benefit appears early and grows with the volume of consumed ground truth. \textit{Therefore, maximizing the volume of labeled samples consumed by the surrogate, as early as possible, maximizes accuracy.}

% \begin{figure}[t]
%     \centering
%     \includegraphics[width=0.8\linewidth]{figures/exp_motivation/fig_single_multicontext_benchmark_ptychonn.pdf}
%     \caption{Per-sample inference latency: single context vs.\ multi-context. \textit{Lower is better.}}
%     \label{fig:single_multicontext_benchmark}
% \end{figure}

\paragraph*{\bf Observation 2: Multi-context concurrency destabilizes inference latency}
We quantify the cost of the obvious way of obtaining online fine-tuning: running two model instances side by side in separate CUDA contexts, one serving and one training. \cref{fig:single_multicontext_benchmark} plots per-sample inference latency vs batch size, with error bars to show fluctuation. A single-context serving-only deployment exhibits low and predictable latency that grows mildly with batch size. The multi-context configuration shows higher mean latency and, more damagingly, enormous spread--the GPU's streaming multiprocessors, the launch queue, and memory bandwidth are contended between the two contexts, and the kernel scheduler non-deterministically interleaves the contexts. \textit{Consequently, the latency variance introduced by multi-context CUDA execution will make latency estimation, and thereby meeting a hard inference deadline challenging.}

\textit{Driven by these two observations}, \proj reconciles the conflict between accuracy and latency stability by running both inference and fine-tuning in a \emph{single} CUDA context.
% =================================================
% =================================================
% =================================================

\section{System Design}
\label{section:design}

Following the observations presented in \cref{section:motivation}, we propose four design principles for \proj, a novel approach for DNN surrogate fine-tuning for inference on deadline-constrained workflows. 
% \MMR{Commenting this sentence as it is used in the Design Principles section below.}
% From the four principles presented below, the first two address the unification of inference and fine-tuning in a single CUDA context, and the last two address the scheduling of the resulting fused batches under a hard deadline.

% TAKEN FROM AI OUTPUT
% =================================================
% =================================================
% =================================================
\subsection{Design Principles}
\label{subsection:sysdesign_design_principle}

\textit{Driven by the two observations of \cref{section:motivation}}, we organize \proj around four design principles. The first two address the unification of inference and fine-tuning in a single CUDA context, and the last two address the scheduling of the resulting fused batches under a per-unit hard deadline.

\begin{figure}[t]
\centering
\includegraphics[width=0.8\linewidth]{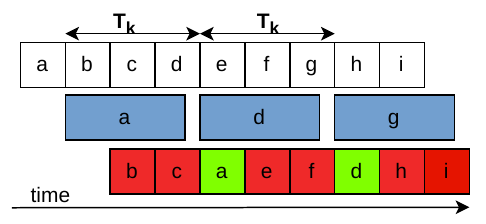}
        \caption{Interleaving of classical ground-truth generation (blue) and DNN surrogate inference (red). Patterns $a, d, g$ are routed to the classical solver, which supply ground truth (green) to surrogate; the surrogate serves the rest.}
    \label{fig:gt_nn_schedule} 
\end{figure}

\paragraph*{\bf P1: Overlap classical ground-truth generation with DNN inference}
We split the stream into intervals $T_k$. Within each interval, the classical solver produces $k$ ground-truth outputs while the DNN surrogate serves the remainder. Each ground-truth output is used twice: directly as the high-fidelity reconstruction for its corresponding unit, and as a labeled training sample for the surrogate. Because the classical solver runs on CPU and the surrogate runs on GPU, the two pipelines overlap naturally with minimal contention. \cref{fig:gt_nn_schedule} illustrates the schedule for $k = 1$: in each interval, pattern $a$ (then $d$, then $g$, $\ldots$) is routed to the classical solver while the remaining patterns are served by the DNN surrogate. The surrogate's replies may arrive out of order, but the downstream reconstruction step reassembles them by index. This design maximizes the supply of fresh ground truth without sacrificing per-unit deadlines on the patterns assigned to the surrogate.

\paragraph*{\bf P2: Fused forward pass with selective backward pass}
The central idea of \proj is to fuse fine-tuning and inference into a single forward pass. Within an iteration, we form a batch that contains both \emph{labeled samples} (diffraction patterns paired with classical ground truth, used for fine-tuning) and \emph{unlabeled samples} (inference requests). The forward pass is run once over the entire batch. We then slice the output: the slice corresponding to the unlabeled samples is dispatched as the inference reply, while the slice corresponding to the labeled samples enters the loss computation and a backward pass that updates the surrogate's parameters. This unification has three benefits. First, it eliminates the redundant forward computation that a naive co-located training/serving setup would incur. Second, because there is only one CUDA context, kernel scheduling is fully sequential, and the contention pathologies of \cref{fig:single_multicontext_benchmark} are reduced. Third, fusing inference requests and labeled samples enables larger and more efficient batches than
% would be possible by
interleaving small inference-only and training-only batches, which would underutilize the GPU. The downside is that a larger fused batch takes longer to complete, and delaying the inference replies it carries. The next principle and the scheduler of \cref{sec:algo} collectively decide how large to make each batch and how to allocate it between labeled (ground truth from the classic solver) and unlabeled samples.

\begin{figure}[t]
    \centering
    \includegraphics[width=0.8\linewidth]{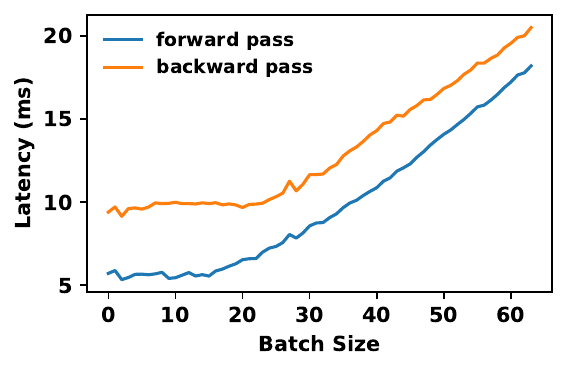}
    \caption{Forward- and backward-pass latency of PtychoNN as a function of batch size. 
    It increases visibly only after saturating the GPU parallelism, which allows batching up to a certain threshold for free \textit{(Lower is better).}}
    \label{fig:benchmarkforwardbackward}
\end{figure}

% \JM{move some part of this and Figure 5 to Section 4 (Motivating observation) and put P3 before P1}
\paragraph*{\bf P3: Exploit the non-linear latency curve}
\cref{fig:benchmarkforwardbackward} shows the forward- and backward-pass latency of PtychoNN as a function of batch size. Both curves are nearly flat up to a model- and hardware-specific knee, and grow approximately linearly thereafter, which makes per-sample inference latency nearly constant as observed in Figure~\ref{fig:single_multicontext_benchmark}. The flat region corresponds to batch sizes that do not saturate the GPU's streaming multiprocessors; the linear region corresponds to batch sizes large enough that the workload is compute-bound. \proj exploits this curve in two ways. Below the knee, adding labeled or unlabeled samples to a batch is essentially free: it does not increase latency but does increase useful work, so a scheduler should never under-fill a batch. Above the knee, the per-sample cost is well-modeled by a linear function. This and the higher stability due to P2 allows the scheduler to predict the latency of an arbitrary $(k_\text{train}, k_\text{infer})$ batch combination ahead of time. Because the model is fixed for the duration of the experiment, this curve is profiled once before deployment.

\paragraph*{\bf P4: Compute the batch composition with a dynamic-programming scheduler}
The intervals $T_k$ are by construction repetitive: each interval brings in the same number of labeled samples and (in expectation) the same number of inference requests. The scheduler's job, then, is to find an ordered sequence of fused batches that processes a single interval's workload while (i)~meeting every inference's per-unit deadline or (ii)~minimizing the interval's makespan while trying to process as much processing of inference units within the deadline. As, per-batch latency is predictable (with higher stability in unified context) from $F(b)$ and $B(b)$ (P3), the per-interval scheduling problem has overlapping subproblems and optimal substructure: the schedule for $i$ remaining labeled samples and $j$ remaining inference requests is determined by the choice of the next batch composition $(k, l)$ together with the schedule for $(i-k, j-l)$. We solve this with dynamic programming, presented formally in \cref{sec:algo}; the schedule is computed once, before the experiment starts.

\subsection{Unified Iteration and Batch Scheduler Algorithm}
\label{sec:algo}

\cref{alg:iteration} specifies how a single iteration is executed in a shared context. Given the number of labeled samples $n_b$ and the number of unlabeled inference samples in the current batch, we run the forward pass over the union, slice off the inference outputs to return them to the consumer, and run the backward pass only over the labeled slice. What remains is to compose batches so that an interval's labeled samples and inference requests are processed under the deadline. The scheduler optimizes a systems proxy for \cref{eq:objective}, not MSE directly: it serves as many requests within $D$ as possible and consumes labeled samples early, since misses fall back to $g(\cdot)$ and Observation~1 ties MSE to consumed ground truth.

\begin{algorithm}[t]
\caption{\proj single-context iteration} 
\label{alg:iteration}
{\small
\KwIn{Pending inference queue $Q_\text{inf}$; pending labeled queue $Q_\text{tr}$; iteration composition $(n_b, n_f)$ with $n_b$ labeled and $n_f - n_b$ inference samples}
$L \leftarrow$ first $n_b$ elements of $Q_\text{tr}$\;
$U \leftarrow$ first $(n_f - n_b)$ elements of $Q_\text{inf}$\;
$O \leftarrow \texttt{nn.forward}(L \cup U)$\;
\textbf{return} the slice of $O$ corresponding to $U$ to the consumer\;
\If{$n_b > 0$}{
  $\ell \leftarrow \texttt{loss}(O[L], \mathrm{labels}(L))$\;
  $\texttt{nn.backward}(\ell)$\;
  $\texttt{optimizer.step}()$\;
}
} % \small
\end{algorithm}

\cref{alg:optimal_makespan_scheduler} specifies the dynamic-programming~(DP) scheduler. We define $dur[i][j]$ as the minimum time at which a schedule can complete after processing the first $i$ labeled samples and the first $j$ inference requests of an interval. The recurrence considers every batch composition $(k, l)$ with $1 \le k+l \le M_{\text{mem}}$ ($M_{\text{mem}}$: the maximum mini-batch size GPU can fit) as the final batch of the partial schedule: that batch consumes the previous $k$ labeled samples and $l$ inference requests, takes iteration time $F(k+l) + B(k)$ (forward over the full batch, backward over the labeled slice), and cannot start before the arrival time of the most recent inference request in the batch. The deadline constraint requires that the iteration complete within $D$ of the \emph{earliest}-arriving inference in the batch (or, if $l = 0$, imposes no deadline constraint from the inference side). The base case is $dur[0][0] = 0$. After the table is filled, the schedule is reconstructed by walking backward from $(N_{\text{tr}}, N_{\text{inf}})$ through the recorded choices $Sc[i][j]$. If $dur[N_{\text{tr}}][N_{\text{inf}}] = \infty$, no feasible schedule exists for the chosen interval size, and the system reduces $N_{\text{tr}}$ (i.e., admits fewer ground-truth samples per interval) until feasibility is restored. The search space includes training-only ($l = 0$) and inference-only ($k = 0$) batches, so a single-context policy that never fuses is a feasible point of it; the returned schedule is no worse under the profiled latency model. \proj fuses only where the DP finds it profitable.

\begin{algorithm}
\caption{Best-effort DP-based batch scheduler} 
\label{alg:optimal_makespan_scheduler}
{\small
\KwIn{Labeled budget $N_{\text{tr}}$; inference budget $N_{\text{inf}}$; inter-arrival time $\Delta = 1/d_{rate}$; deadline $D$; mini-batch cap $M_{\text{mem}}$; latency profiles $F(\cdot), B(\cdot)$}
\KwOut{Schedule $S$: ordered list of $(k, l)$ pairs}
$dur[i][j] \leftarrow \infty$ for all $i, j$;\quad $dur[0][0] \leftarrow 0$\;
$Sc[i][j] \leftarrow (0, 0)$ for all $i, j$\;
\For{$i = 0$ \KwTo $N_{\text{tr}}$}{
  \For{$j = 0$ \KwTo $N_{\text{inf}}$}{
    \If{$i = 0$ \textbf{and} $j = 0$}{\textbf{continue}}
    \For{$k = 0$ \KwTo $\min(M_{\text{mem}}, i)$}{
      \For{$l = 0$ \KwTo $\min(M_{\text{mem}} - k, j)$}{
        \If{$k + l = 0$ \textbf{or} $dur[i-k][j-l] = \infty$}{\textbf{continue}}
        $t_\text{iter} \leftarrow F(k+l) + B(k)$\;
        $t_\text{arr} \leftarrow j \cdot \Delta$ \tcp*{j-th infer.\ request arrival}
        $t_\text{start} \leftarrow \max(dur[i-k][j-l],\, t_\text{arr})$\;
        $t_\text{fin} \leftarrow t_\text{start} + t_\text{iter}$\;
        \If(\tcp*[h]{deadline of earliest infer.\ in batch}){$l > 0$ \textbf{and} $t_\text{fin} > (j-l+1) \cdot \Delta + D$}{\textbf{continue}}
        \If{$t_\text{fin} < dur[i][j]$}{
          $dur[i][j] \leftarrow t_\text{fin}$\;
          $Sc[i][j] \leftarrow (k, l)$\;
        }
      }
    }
  }
}
\If{$dur[N_{\text{tr}}][N_{\text{inf}}] = \infty$}{\textbf{return} \textsc{Infeasible}}
$S \leftarrow [\,]$;\quad $(i, j) \leftarrow (N_{\text{tr}}, N_{\text{inf}})$\;
\While{$(i, j) \neq (0, 0)$}{
  $(k, l) \leftarrow Sc[i][j]$;\quad \textbf{prepend} $(k, l)$ to $S$;\quad $(i, j) \leftarrow (i-k, j-l)$\;
}
\textbf{return} $S$\;
} % \small
\end{algorithm}

The scheduler's time complexity is $O(N_{\text{tr}} \cdot N_{\text{inf}} \cdot M_{\text{mem}}^2)$, which is polynomial in the workload size and dominated by the $M_{\text{mem}}^2$ factor in practice. The quadratic term is device-bound: a weaker GPU admits a smaller $M_{\text{mem}}$ and a cheaper search. For our PtychoNN configuration, the scheduler returns in under a second, and, because the interval is repetitive, it runs exactly once at the start of the experiment. Larger GPU memory and smaller models raise $M_{\text{mem}}$; in our deployment, $N_{\text{tr}} = 1$ per interval (one ground-truth sample per interval, matching the classical solver throughput) and $N_{\text{inf}}$ scales with $d_{rate}$.

\begin{figure*}[t]
    \centering
    \includegraphics[width=0.8\linewidth]{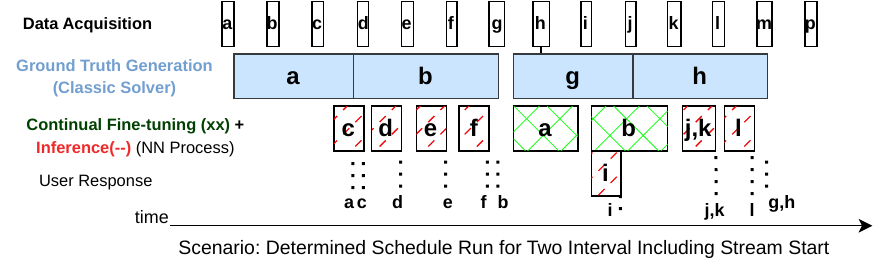}
    \caption{Two consecutive intervals of an example \proj schedule with $N_{\text{tr}} = 2$. In the first interval (no ground truth yet) the surrogate runs inference-only batches. In the second interval, the scheduler first fine-tunes on labeled sample $a$, then runs a fused batch $\{b, i\}$, then a fused batch $\{j, k\}$, then a final batch $\{l\}$. Box width represents processing latency for the corresponding samples.  Notice, $\{a\}$, $\{b\}$ with hatched background indicates forward + backward pass.}
    \label{fig:unipipe_full_example} 
\end{figure*}

\cref{fig:unipipe_full_example} walks through a representative two-interval schedule produced by \cref{alg:optimal_makespan_scheduler}. In the first interval, no labeled samples are yet available, so the scheduler defaults to inference-only batches. In the second interval, the scheduler first issues a labeled-only fine-tuning batch on $a$ (this is affordable because the next inference $i$ has enough slack), then a fused batch $\{b, i\}$ that simultaneously fine-tunes on $b$ and replies to $i$, then a fused batch $\{j, k\}$ in which two inferences share one forward pass (no labeled samples remain), and finally an inference-only batch on $l$. Batching $\{j, k, l\}$ together would have been feasible, but the scheduler prefers to reply to $j$ and $k$ sooner because doing so does not violate $l$'s deadline.

\subsection{Implementation}
\label{subsection:unipipe_implementation}

We implement \proj as three coordinated Python processes. A \emph{streaming process} simulates the sensor by emitting diffraction patterns at rate $d_{rate}$ and dropping any pattern whose deadline $D$ has expired. A  \emph{classical solver process} consumes patterns at its native rate $c_{rate}$, produces high-fidelity outputs, and writes them as labeled examples to a shared-memory queue (\texttt{/dev/shm}). A \emph{DNN surrogate process} executes \cref{alg:iteration} according to the schedule produced by \cref{alg:optimal_makespan_scheduler}, reading inference requests and labeled samples from their respective queues. The surrogate is the open-source PtychoNN architecture\footnote{\url{https://github.com/mcherukara/PtychoNN}}, whose forward pass is mode-invariant (no batch-norm or dropout layers), so fusing labeled and unlabeled samples in the same forward pass is straightforward; models with mode-dependent layers would require running the corresponding sub-batches in their appropriate mode. The fused output tensor is sliced, the loss is computed from the labeled slice, and backward is invoked on that loss, so unlabeled outputs do not affect the update. Autograd still retains the whole batch's activations, since the forward pass is one graph, so an iteration costs the memory of a full-batch training step, which $M_{\text{mem}}$ bounds. Missed inferences are filled in by interpolation from the neighboring reconstructed patches, an operation that takes microseconds and overlaps with ongoing processing. The implementation, including the dynamic-programming scheduler, the three-process runtime, and the PtychoNN integration, is publicly available\footnote{\url{https://github.com/Jaiaid/unipipe\_ptychonn\_experiment}}.

% =================================================
% =================================================
% =================================================
\section{Experimental Results}
\label{section:evaluation}

% \subsection{Methodology}
\subsection{Evaluation Methodology}
\label{sec:method}

\subsubsection{\bf Evaluation Testbed}
\label{subsubsection:eval_env}
Our evaluations are conducted on a single-GPU server comprising RTX A4000-16GB GPU, Intel(R) Xeon(R) Gold 5315Y CPU (8 physical cores, 16 threads), and 512~GB of DRAM, connected via PCIe Gen~3 to the GPU. As noted in various studies~\cite{chuang2023implementation,sinha2024towards,hoang2024evaluating}, our system is representative of a real-world edge deployment. Our software stack is composed of Ubuntu 23.04, CUDA version 12.2, Python 3.11, and PyTorch 2.6.0.

\subsubsection{\bf Use Case: Real-world Ptychography Application}
Throughout our evaluations, we consider a real-world use-case as described in \S\ref{section:prob_formulation}. 
% The classical computations are done using the ePIE approach~\cite{rodenburg2007transmission,maiden2009improved}, which leverages the CPU's multi-threading capabilities to process diffraction images at 16 samples/second. 
The DNN surrogate is PtychoNN~\cite{cherukara2020ai}, $\sim4.7$M parameters, serving 500--3500 inferences/second and fine-tuning at 70--1700 samples/second depending on batch size~(\S\ref{subsection:sysdesign_design_principle}). Given its large and frequent matrix operations, both training and inference run exclusively on the GPU.

\subsubsection{\bf Dataset}
We use two datasets, differing in number of scans and individual scan size: \textbf{Dataset 1} for system performance evaluation, and both for inference quality.

\paragraph{\bf Dataset 1}
Raster scans of $161\times161\approx$26K diffraction patterns from the PtychoNN repository~\cite{cherukara2020ai}, collectively covering a single $1800nm\times1800nm$ object. Each sample is a triplet of the raw far-field diffraction scan and the corresponding real-space amplitude and phase images.

\paragraph{\bf Dataset 2}
This dataset contains raster scans of $186\times186\approx$35k scans collected at APS~\cite{kerby2023advanced} by scanning a calibration plate containing a random etched pattern in $1.5\mu m$ tungsten~\cite{du2024predicting}.

% TAKEN FROM AI OUTPUT
% =================================================
% =================================================
% =================================================
% \paragraph*{\bf Compared approaches}
\subsubsection{\bf Compared Approaches}
We compare \proj against two baselines that span the prevailing design space.

\paragraph{\bf Pretrained}
The PtychoNN surrogate is trained offline on $\sim\!20\%$ of the dataset and frozen for the live run. This baseline represents the highest serving throughput and the lowest accuracy ceiling, since no in-situ adaptation occurs.

\paragraph{\bf Multi-Context}
Two surrogate instances colocate on the GPU in separate CUDA contexts, following the typical design of recent online-adaptation systems~\cite{wang2019adaptive,wortsman2022model,houlsby2019parameter}. One instance serves inference; the other fine-tunes; the serving instance periodically pulls updated weights from the fine-tuning instance. This baseline closes the accuracy gap of \emph{Pretrained} but pays the contention cost characterized in Observation~2.

\paragraph{\bf \proj (Our Approach)}
The single-context system of \cref{section:design}, which fuses inference requests and labeled samples in one forward pass scheduled by \cref{alg:optimal_makespan_scheduler}. This configuration aims to retain the latency stability of a serving-only deployment while fine-tuning on ground truth.

% \paragraph*{\bf Parameters and metrics}
\subsubsection{\bf Parameters and Metrics}
We sweep $d_{rate}$ from 1,000 to 5,000 patterns per second in steps of 1,000, spanning the regime where modern detectors operate up to the rates expected for next-generation beamlines (for $512 \times 512$ patterns with 4-byte values, this corresponds to roughly 8--40~Gbps). The classical solver throughput $c_{rate}$ is fixed at 16 samples per second, consistent with reported edge-CPU rates for iterative Ptychographic reconstruction on a small region of interest~\cite{babu2023deep}. We set $D = 80$~ms, the smallest round value above the solver's per-sample latency $1/c_{rate} = 62.5$~ms; below that bound, a classically reconstructed unit would miss its own deadline, leaving the solver's output usable only as a training label. Each experiment is repeated 6 times; we report the mean, minimum, and maximum for each experiment. We report two primary metrics: \emph{inference accuracy}, measured as the mean squared error (MSE) of the surrogate's phase-image output against the classical reconstruction, and \emph{miss rate}, the fraction of inferences that fail to complete within deadline $D$.

% \paragraph*{\bf Evaluation methodology}
We organize the evaluation around four sweeps that progressively isolate the contributions of \proj's design, and close with an ablation attributing the improvement to individual design principles.

\subsection{Inference Quality}
\label{subsubsection:exp_eval_infer_quality}

\begin{figure}[t]
    \centering
    \includegraphics[width=\linewidth]{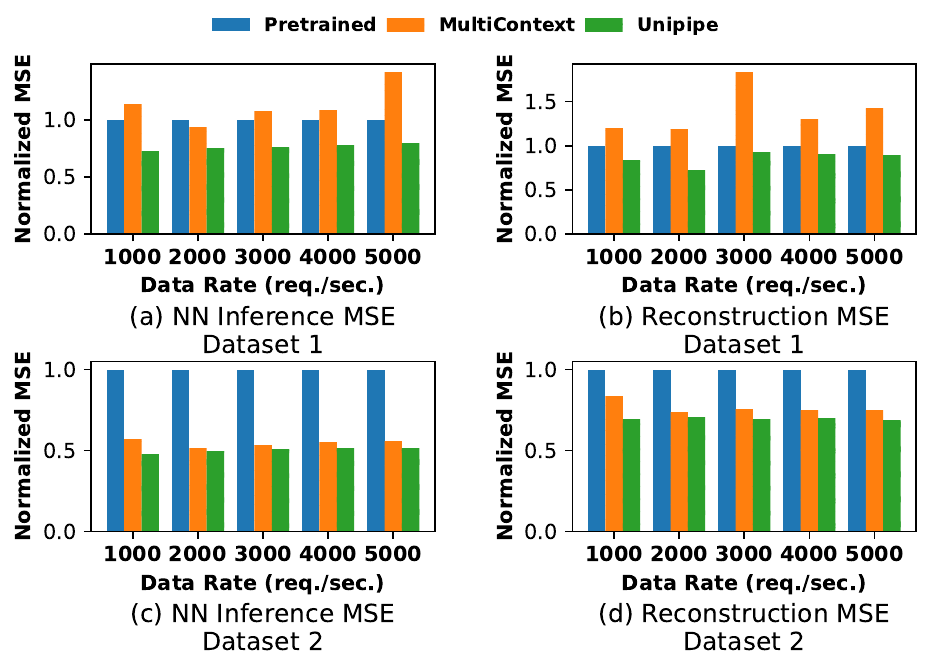}
    \caption{Inference and reconstruction MSE across data arrival rates. (a, c) MSE of the DNN-surrogate-served samples only. (b, d) MSE of the final reconstructed image, which combines surrogate outputs, classical outputs, and neighbor-interpolated fallbacks for missed inferences. (a, b) Dataset~1; (c, d) Dataset~2. \textit{Lower is better.}}
    \label{fig:eval_mse_metric}
\end{figure}

\cref{fig:eval_mse_metric}(a) and \cref{fig:eval_mse_metric}(c) report the per-sample inference MSE of the DNN-served subset of patterns across data arrival rates. \proj outperforms both baselines at every $d_{rate}$, achieving at least $\sim\!30\%$ and up to $\sim\!50\%$ lower inference MSE than the Pretrained baseline. The Pretrained baseline cannot improve over its offline-trained accuracy; the Multi-Context baseline does fine-tune online, but its high miss rate (\cref{subsubsection:exp_eval_performance_missrate}) reduces the volume of inferences that benefit from the fine-tuned model. \cref{fig:eval_mse_metric}(b) and \cref{fig:eval_mse_metric}(d) report the holistic reconstruction MSE, which combines DNN-surrogate outputs, classical outputs, and neighbor-interpolated fallbacks. \proj retains the end-to-end accuracy advantage, achieving at least $\sim\!15\%$ and up to $\sim\!30\%$ lower reconstruction MSE than the Pretrained approach. The smaller relative gap on reconstruction MSE compared to inference MSE reflects the fact that classical outputs and interpolation contribute fixed components to the aggregate error regardless of the approach used for the DNN-served patterns. The two panel pairs thus separate the mechanisms: (a, c) isolate the gain from consuming labeled samples earlier, (b, d) add the gain from fewer misses. \textit{This 
% demonstrates empirically 
shows
that \proj's unified context inference and fine-tuning can improve the accuracy metric.}

\subsection{Throughput and Miss Rate}
\label{subsubsection:exp_eval_performance_missrate}

\begin{figure*}[t]    
    \centering
    \includegraphics[width=\linewidth]{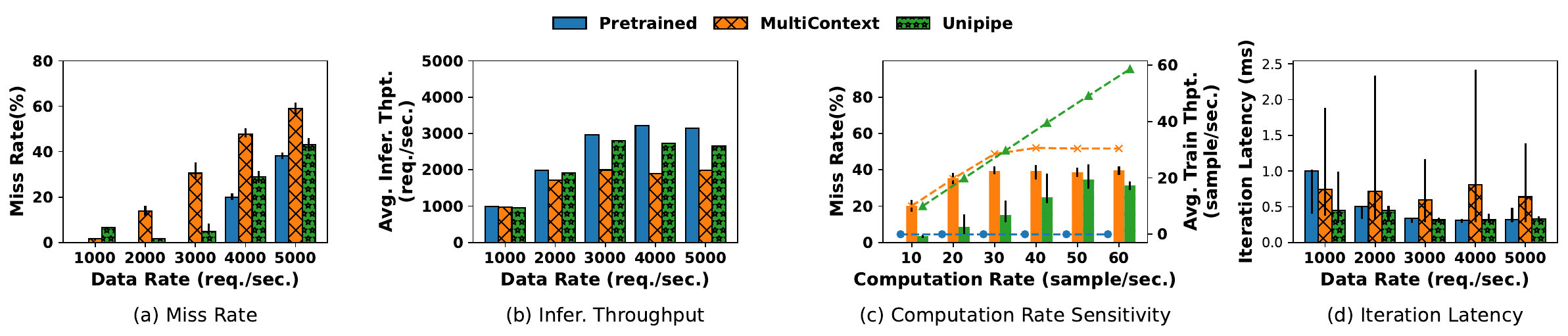}
    \caption{System performance across the three approaches. \textbf{(a)} Miss rate vs.\ data arrival rate (\textit{Lower is better}). \textbf{(b)} Avg. inference throughput vs.\ data arrival rate (\textit{Higher is better}). \textbf{(c)} Miss rate at different $c_{rate}$ and training throughput (\textit{Lower is better}). \textbf{(d)} Per-sample iteration latency: mean and 95\textsuperscript{th}-percentile (\textit{Lower is better}).}
    \label{fig:eval_system_metric} 
    \vspace{-15pt}
\end{figure*}

\cref{fig:eval_system_metric}(a) reports miss rate as a function of $d_{rate}$. \proj tracks the Pretrained baseline closely across the entire sweep despite also processing training data, whereas Multi-Context suffers a $13\%-22\%$ higher miss rate at $d_{rate} \ge 2000$ patterns/sec. (PtychoNN infers at $\sim3000$ patterns/sec.), and the gap widens monotonically with the arrival rate. The cause is the latency variability of \cref{fig:single_multicontext_benchmark}: under contention, Multi-Context's fluctuating per-batch latency causes frequent deadline violations. The pattern persists up to 10000 samples/sec. (i.e., 80~Gbps), though the gap narrows and accuracy drops. \textit{Consequently, \proj's single-context design closely attains the deadline-meeting behavior of a serving-only deployment while still benefiting from continual fine-tuning.}

\cref{fig:eval_system_metric}(b) shows net inference throughput. 
% At the lowest arrival rate the GPU is underutilized and all three approaches keep up. 
% The 
GPU is underutilized at the lowest arrival rate and all three approaches keep up. 
Pretrained leads at higher rates, as expected of an inference-only deployment; \proj follows within a small margin, while Multi-Context underperforms because of the latency fluctuation of \cref{section:motivation}. \cref{fig:eval_system_metric}(c) reports miss rate across varying classical-solver throughput $c_{rate}$, i.e., across a growing supply of training data. The gap between Multi-Context and \proj shrinks: Multi-Context keeps up while the training set is small, but its throughput plateaus as $c_{rate}$ grows, processing only $50\%$ of the training data \proj does, while \proj averages a better miss rate (its slightly worse maximum at $c_{rate}=50$ samples/sec. stems from one outlier among six runs). \textit{This shows that \proj approaches the inference-only throughput ceiling while delivering better accuracy and 
% can sustain 
sustains
higher total throughput given the same hardware resource, and achieving a trade-off between accuracy (\S\ref{subsubsection:exp_eval_infer_quality}) and Miss-rate that the two baselines cannot match.}

\subsection{Latency Stability}
\label{subsubsection:exp_eval_performance_fwpass_latency}

\cref{fig:eval_system_metric}(d) reports average and 95\textsuperscript{th}-percentile per-sample iteration latency. We use the 95\textsuperscript{th} percentile instead of the absolute maximum to filter warm-up outliers. \proj's mean per-sample latency is lower than that of the Multi-Context baseline. On top of that its 95\textsuperscript{th}-percentile latency is at most $57\%$ above its mean (smaller mean), compared to at most $106\%$ larger than the Multi-Context's mean (which is larger than \proj mean to begin with), i.e., the single-context fused-batch design reduces the relative latency spread by nearly half. \textit{Thereby confirming the qualitative diagnosis of Observation~2}, this is precisely the property the scheduler of \cref{alg:optimal_makespan_scheduler} needs to make safe deadline predictions. The residual variability produces the small remaining miss-rate gap to a hypothetical zero-jitter system.

\subsection{Effect of Static Resource Partitioning via MPS}
\label{subsubsection:exp_MPS}

\begin{figure}[t]
    \centering
    \begin{minipage}[t]{0.49\linewidth}
        \includegraphics[width=\linewidth]{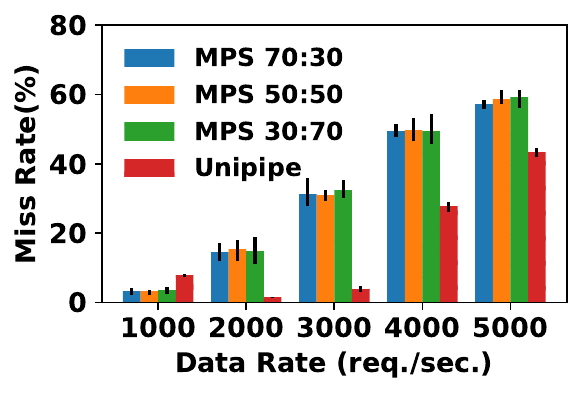}
        \caption{MPS-partitioned Multi-Context vs Unipipe. Inference to training ratios
        are marked in the legend.}
        \label{fig:exp_mpsstatic_split_vs_unipipe}
    \end{minipage}\hfill
    \begin{minipage}[t]{0.49\linewidth}
        \includegraphics[width=\linewidth]{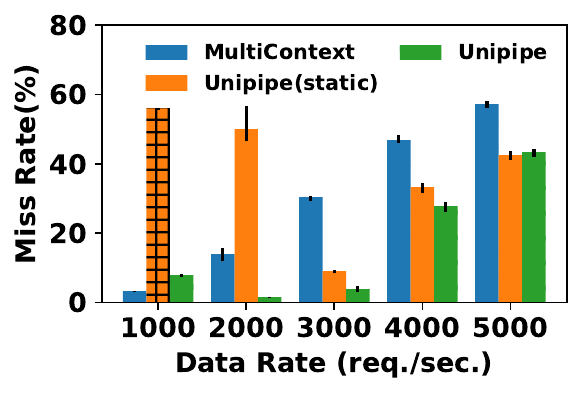}
        \caption{Comparison of contribution in miss rate reduction of each \proj design principle to the miss rate.}
        \label{fig:exp_ablation}
    \end{minipage}
\end{figure}

The Multi-Context baseline of \cref{fig:eval_system_metric} relies on NVIDIA's default time-sliced scheduling between contexts. A reasonable hypothesis is that a manually tuned static partition of GPU streaming multiprocessors between the inference and training contexts, via NVIDIA's Multi-Process Service (MPS)~\cite{cuda_mps,cuda_mps_oakridge}, could close the gap to \proj by trading throughput for predictability. \cref{fig:exp_mpsstatic_split_vs_unipipe} compares \proj against Multi-Context configurations under several MPS splits (inference:training share of SMs). Across every split and across every non-trivial $d_{rate}$, \proj posts a lower miss rate, with the gap reaching up to $\sim\!18\%$. We also evaluate for 90:10 and 10:90 splits, which show a similar pattern of no significant difference in Miss rate in Multi-Context (not shown for figure clarity and space constraint). \textit{Thereby demonstrating that} static partitioning cannot recover the latency-stability benefit of single-context execution, because contention also occurs at the global memory and the kernel-launch path, neither of which is handled by MPS partitions~\cite{villarrubia2026comprehensive}.

\subsection{Ablation Study}
\label{subsubsection:exp_ablation}

\cref{fig:exp_ablation} attributes \proj's improvement to its individual design principles. We compare three configurations: (i)~\textbf{Multi-Context}, the baseline; (ii)~\textbf{\proj (static)}, which adopts the single-context fused-batch design (P1, P2) but uses a fixed batch size pinned at the GPU-saturation knee rather than running the scheduler (omits P3, P4); and (iii)~\textbf{\proj}, the full system. The single-context fused-batch design alone (P1, P2) captures the bulk of the miss-rate improvement at high data rates ($d_{rate} \ge 3,000$), where contention dominates in Multi-Context. The scheduler (P3, P4) matters most at low rates: with the GPU under-pressured, a static batch size yields too many small iterations, whereas the scheduler picks larger fused batches that amortize per-iteration overhead. At high rates the scheduler converges on a batch size near the saturation knee, which is why the two \proj variants converge there. \textit{This means that the unification principles (P1, P2) and the scheduling principles (P3, P4) are complementary: each dominates a different region of the operating envelope.}

% =================================================
% =================================================
% =================================================
\section{Conclusions}

We propose \proj, a single-context system unifying inference serving and fine-tuning by combining: overlapped classical ground-truth generation and DNN inference; a fused forward pass over labeled and unlabeled samples with a selective backward pass on the labeled subset; exploitation of the non-linear per-sample latency curve of modern GPUs; and a polynomial-time dynamic-programming scheduler that computes the batch composition under the deadline. On a real-world Ptychographic workflow with PtychoNN, \proj lowers reconstruction MSE by up to $30\%$ and inference MSE by up to $50\%$ over a frozen pretrained surrogate, and reduces the deadline-miss rate by $13\%-22\%$ while shrinking the 95\textsuperscript{th}-percentile latency from $106\%$ to $57\%$ of the mean over a multi-context baseline.

In the future, we plan to extend \proj to adaptive selection of the labeled training data in planning the schedule to achieve better accuracy and miss rate, and evaluate other surrogate architectures, e.g., transformers. We also plan to evaluate \proj for workflows other than Ptychography.

% following instruction from 
% https://www.escience-conference.org/2025/call-for-papers
\section{Acknowledgements}
% We acknowledge use of generative AI to polish (rephrase, grammar, and spelling check) the paper after we manually write the draft. We use Claude opus through OpenCode for the generation. We declare that the generated results and experiments are manually thought upon and performed. We assume responsibility for the presented results.

We acknowledge the use of AI tools, particularly Claude Opus 4.7, to improve the flow and grammar of the paper. The authors have verified and take full responsibility of text across all sections of the paper. 
This work is supported in part by 
the U.S. Department of Energy (DOE), Office of Advanced Scientific Computing Research (ASCR) under contract
0F–60169
and the National Science Foundation (NSF) under Awards No. 2411386/2411387.

% =================================================
% =================================================
% =================================================
\balance

\bibliographystyle{IEEEtran}
\bibliography{refdb}

\end{document}